\documentclass[runningheads]{llncs}
\usepackage{graphicx}
\begin{document}
\title{Improving Amharic Handwritten Word Recognition Using Auxiliary Task}
%
%
\author{Mesay Samuel Gondere\inst{1} \and
Lars Schmidt-Thieme\inst{2} \and
Durga Prasad Sharma\inst{3} \and
Abiot Sinamo Boltena\inst{4}}
\authorrunning{M. Samuel et al.}
%
\institute{Arba Minch University, Faculty of Computing and Software Engineering, Ethiopia
\email{mesay.samuel@amu.edu.et} \and
Information Systems and Machine Learning Lab, 31141 Hildesheim, Germany
\email{schmidt-thieme@ismll.uni-hildesheim.de} \and
AMUIT MOEFDRE under UNDP
\email{dp.shiv08@gmail.com} \and
Ministry of Innovation and Technology, Federal Democratic Republic of Ethiopia
\email{abiotsinamo35@gmail.com}}

\maketitle              
\begin{abstract}
Amharic is one of the official languages of the Federal Democratic Republic of Ethiopia. It is one of the languages that use an Ethiopic script which is derived from Gee'z, ancient and currently a liturgical language. Amharic is also one of the most widely used literature-rich languages of Ethiopia. There are very limited innovative and customized research works in Amharic optical character recognition (OCR) in general and Amharic handwritten text recognition in particular. In this study, Amharic handwritten word recognition will be investigated. State-of-the-art deep learning techniques including convolutional neural networks together with recurrent neural networks and connectionist temporal classification (CTC) loss were used to make the recognition in an end-to-end fashion. More importantly, an innovative way of complementing the loss function using the auxiliary task from the row-wise similarities of the Amharic alphabet was tested to show a significant recognition improvement over a baseline method. Such findings will promote innovative problem-specific solutions as well as will open insight to a generalized solution that emerges from problem-specific domains.

\keywords{Convolutional Recurrent Neural Networks  \and Handwritten Word Recognition \and Auxiliary Task \and Amharic Handwritten Recognition}
\end{abstract}
\section{Introduction}
Amharic is the official language of the federal government of Ethiopia and other regional states in Ethiopia. It is one of the languages that use an Ethiopic script which is derived from Gee'z, an ancient and currently a liturgical language. Amharic script took all of the symbols in Gee'z and added some new ones that represent sounds not found in Gee'z. Amharic is one of the most widely used literature-rich languages of Ethiopia. It is also the second widely spoken Semitic language in the world after Arabic. There are many highly relevant printed as well as handwritten documents available in Ethiopia. These documents are primarily written in Gee'z and Amharic languages covering a large variety of subjects including religion, history, governance, medicine, philosophy, astronomy, and the like \cite{belay2019amharic,gondere2019handwritten,abebe2013biological}. Hence, due to the need to explore and share the knowledge in these languages, there are further established educational programs outside Ethiopia as well, including Germany, the USA, and China recently. Also nowadays, Amharic document processing and preservation is given much attention by many researchers from the fields of computing, linguistics, and social science \cite{belay2020amharic}.

Far from the fact that rich written document resources in Gee'z and Amharic, the technological and research advancements are limited. Especially innovative and customized research works supporting these languages are very limited. Except from language model perspective, technically addressing Amharic written documents will address Gee'z documents since the Amharic alphabet extends the Geez alphabet. Even though optical character recognition (OCR) for the unconstrained handwritten documents itself is still an open problem, Amharic OCR in general and Amharic handwritten text recognition, in particular, are not widely studied. There are some research works growing with the introduction and potential of recent deep learning techniques addressing OCR problems. However, those works focus on directly adapting off-the-shelf existing methods for the case of Amharic scripts. That is most of the studies are only adapting state-of-the-art techniques which are mainly designed for Latin scripts. Even though such studies will have their own benefit by showing how universally proposed solutions will fit specific problem domains, on the other hand however will hinder innovations that could emerge because of specific problems. In this regard, while contextualized and innovative techniques to address Amharic script specifically is very important, it is overlooked in emerging research works \cite{abdurahman2021ahwr,belay2021blended,yohannes2021amharic}.

Offline handwritten text recognition is the hardest of OCR problems. It is offline because the document is taken or scanned separately beforehand without anticipating any recognition technology. Accordingly, one cannot get or extract any relevant feature during the writing process, unlike online OCR. Another issue is the challenge with the complexity and varied handwriting style of people which is not the case with printed documents. Recent OCR problems are addressed using deep learning techniques and could be either character based, word based, or sequence based. Character based methods focus on finding specific locations of individual characters and recognizing them. Whereas word based methods solve text recognition as a word classification problem, where classes are common words in a specific language. The current state-of-the-art methods use sequence to sequence methods. These methods treat OCR as a sequence labeling problem \cite{jaderberg2014synthetic}.

\begin{figure}[!ht]
	\centering
	\includegraphics[width=0.5\textwidth]{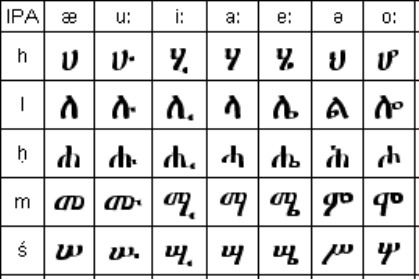}
	\caption{Parts of Amharic alphabet. Source: Omniglot.} \label{amh}
\end{figure}

\subsection{Related Works}
It is worth mentioning some related works and notable attempts in OCR in general and Amharic OCR in particular. A notable and innovative contribution by Assabie et al. \cite{assabie2009hmm} for Amharic word recognition is one of the earlier works. In this paper, writer-independent HMM-based Amharic word recognition for offline handwritten text is presented. The underlying units of the recognition system are a set of primitive strokes whose combinations form handwritten Ethiopic characters. Similar to recent sequence based deep learning based methods, in this paper the recognition phase does not require segmentation of characters \cite{assabie2009hmm,assabie2011offline}. Other recent works have adapted deep learning methods for character based recognition \cite{gondere2019handwritten,belay2018amharic} and sequence based recognition \cite{abdurahman2021ahwr,belay2020amharic}. Sequence based methods particularly address the problem of OCR in an end-to-end fashion using a convolutional neural network as feature extractor from the text image, recurrent neural network as sequence learner, and connectionist temporal classification as a loss function and transcriber. The combination of convolutional neural network (CNN) and recurrent neural network (RNN) is termed as Recurrent Convolution Neural Network (CRNN). A very recent work Abdurahman et al. \cite{abdurahman2021ahwr} designed a custom CNN model and investigated different state-of-the-art CNN models and make available the first public Amharic handwritten word image dataset called HARD-I. The authors also conducted extensive experiments to evaluate the performance of four CRNN-CTC based recognition models by analyzing different combinations of CNN and RNN network architectures. In this paper, the state-of-the-art recognition accuracy for handwritten Amharic words is reported. Similarly, Belay et al. \cite{belay2020amharic} presented an end-to-end Amharic OCR for printed documents.

Apart from adapting off-the-shelf deep learning techniques, there are some innovative works proposing solutions for the problems that arise from deep learning techniques like computational cost and large labeled dataset requirements. Puigcerver \cite{puigcerver2017multidimensional} proposed the usage of only one dimensional RNN and data augmentation to help better recognition and faster computation. Looking for ways that can make improvement through the process of the recognition pipeline is also important. Yousefi et al. \cite{yousefi2015binarization} proposed to skip the binarization step in the OCR pipeline by directly training a 1D Long Short Term Memory (LSTM) network on gray-level text lines for binarization-free OCR in historical documents. Transfer learning is another important technique that enables the transfer of learned artifacts from one problem to solve another problem. This is particularly relevant during a shortage of datasets for the intended problem \cite{jaramillo2018boosting,granet2018transfer}. Granet et al. \cite{granet2018transfer} dealt with transfer learning from heterogeneous datasets with a ground-truth and sharing common properties with a new dataset that has no ground-truth to solve handwriting recognition on historical documents. Jaramillo et al. \cite{jaramillo2018boosting} proposes the boosting of handwriting text recognition in small databases with transfer learning by retraining a fixed part of a huge network. Wang and Hu \cite{wang2017gated} proposed a new architecture named Gated RCNN (GRCNN) for solving OCR by adding a gate to the Recurrent Convolution Layer (RCL), the critical component of RCNN. The gate controls the context modulation in RCL and balances the feed-forward information
and the recurrent information. Finally, relevant input to this study is explored in the work of Gondere et al. \cite{gondere2019handwritten} on how auxiliary tasks are extracted from the Amharic alphabet and improve Amharic character recognition using muli-task learning. A related and typical example of how such a way of addressing specific problem help in innovative and generalized problem solution for multi-script handwritten digit recognition is demonstrated in \cite{gondere2021multiscript}. 

In this study, Amharic handwritten word recognition will be investigated in an end-to-end fashion using CRNN and CTC loss. The word level image datasets were built from character level image datasets of the Amharic alphabet by referring to compiled Amharic words extracted from news and concatenating the character images using writer id. In this study, the usual CRNN model will serve as a baseline model and the study presents an innovative way of complementing the loss function using the auxiliary task from the row-wise similarities of the Amharic alphabet as a proposed model. As can be seen in Figure~\ref{amh} showing part of the Amharic alphabet, the row-wise similarities are exploited as auxiliary classification task to show the significant recognition improvement than a baseline method. The contribution of this study can be summarized in two folds: i) showing an easy and meaningful way of organizing datasets for Amharic handwritten texts, ii) demonstrating an innovative way of improving Amharic handwritten word recognition using an auxiliary task. Finally, the methods and materials used for the study are presented in the following section. Experimental results with the discussions are covered in section three and conclusions are forwarded in the last section.

\section{Methods and Materials}
\subsection{Datasets Preparation}
\begin{figure}[!ht]
	\centering
	\includegraphics[width=\textwidth]{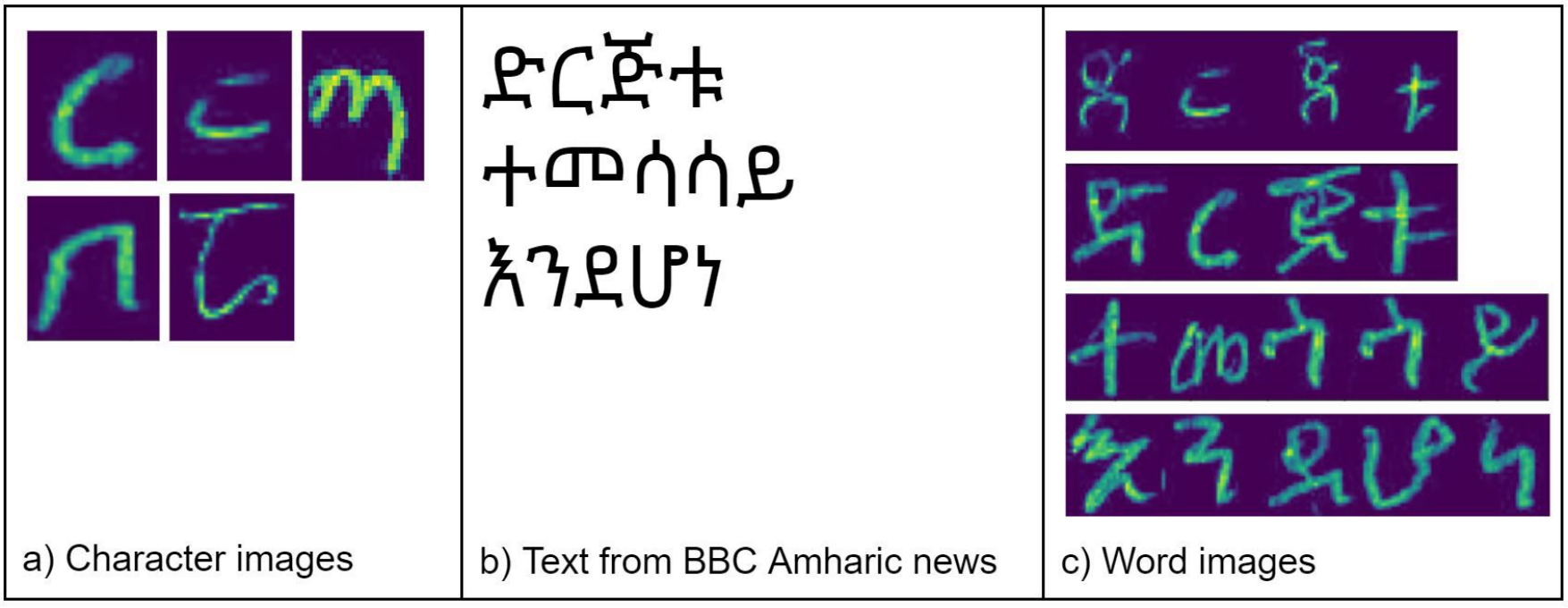}
	\caption{Organization of Amharic handwritten word dataset.} \label{dataset}
\end{figure}

One of the challenges that make the effort of Amharic OCR research work limited is the unavailability of datasets. Very recent and parallel work by Abdurahman et al. \cite{abdurahman2021ahwr} currently made a great contribution by making the Amharic handwritten text dataset available. However, In this study, first, the dataset for Amharic handwritten characters was organized from Assabie et al. \cite{assabie2009comprehensive} and Gondere et al. \cite{gondere2019handwritten}. It is organized to 265 Amharic characters by 100,10 and 10 three groups of separate writers for training, validation, and test sets respectively. Each character image has its label, row label, and writer id. Hence, the row label is used for training the auxiliary task and the writer id is used to generate word images from the same writer by concatenating its character images and referring to the compiled Amharic words extracted from the BBC Amharic news website. Throughout the process of organizing the datasets, python scripts are written and used. Figure~\ref{dataset} shows how the dataset organization was done.

In such data shortage scenarios, it is important to create a reasonable way of organizing datasets to make machine learning experiments possible. Accordingly, the usage of electronic texts and the corresponding character image from the alphabet written by different writers to generate word images allows the inclusion of a large number of characters and a varied natural sequence of characters due to unrestricted coverage of real-world electronic texts. This method is manageable to get a large number of writers while getting only a fixed number of Amharic characters written by respondents. More importantly, the sequence learning method employed in this study and the nature of Amharic writing which is not cursive makes this dataset organization feasible and reasonable. Hence in this study, 256,100 word images are used for training set as 2561 words written by each of the 100 writers. Likewise, two sets of 3,200 word images are used as 320 words written by each of the 10 writers for validation and test sets. The minimum and maximum word lengths are two and twelve respectively and 223 characters were incorporated out of the 265 Amharic characters. 

\begin{figure}[!ht]
	\centering
	\includegraphics[width=\textwidth]{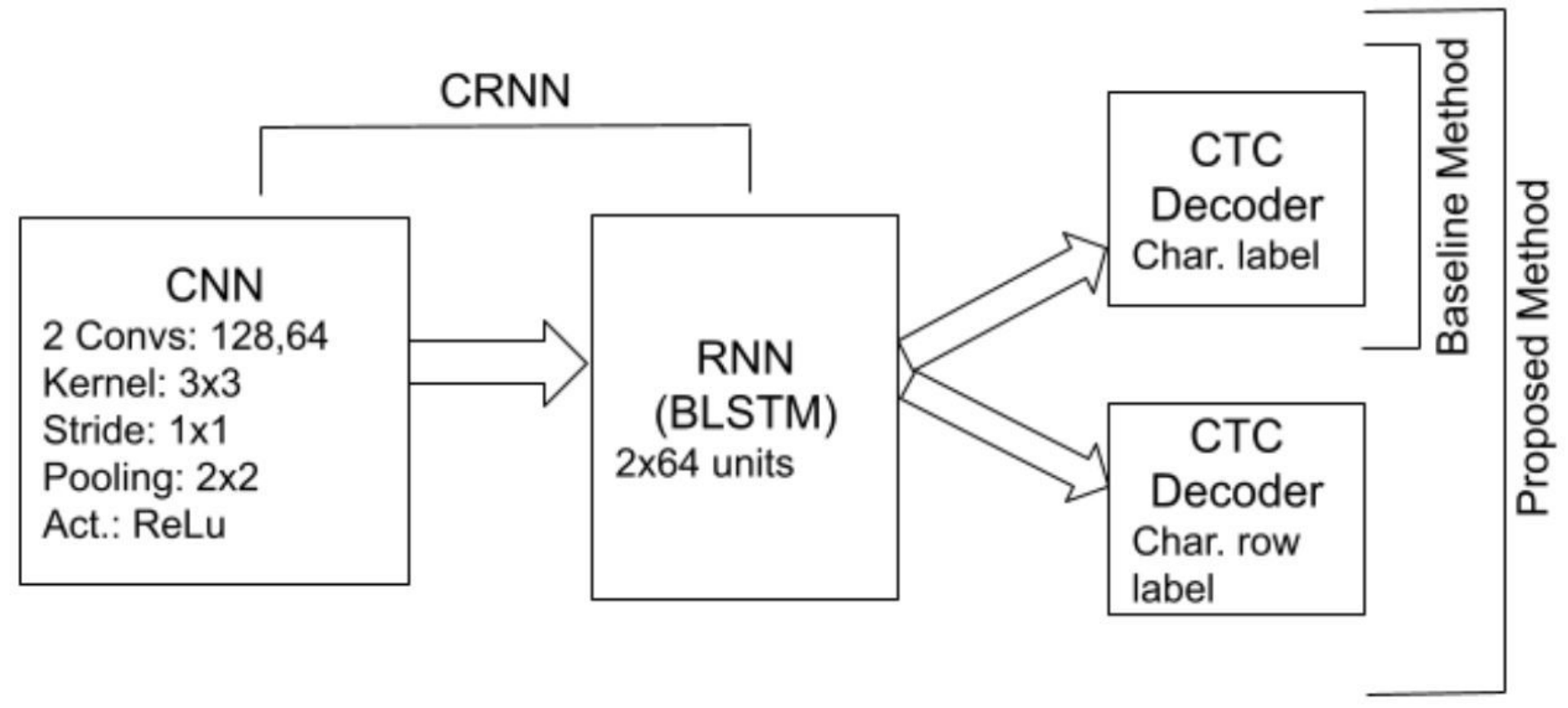}
	\caption{Architecture of the baseline and proposed models} \label{model_architecture}
\end{figure}

Connectionist temporal classification (CTC) is the key to avoiding segmentation at a character level, which greatly facilitates the labeling task \cite{jaramillo2018boosting}. Accordingly, recent works employ CRNN with CTC loss in an end-to-end fashion \cite{belay2020amharic,abdurahman2021ahwr}. The deep CNN allows strong and high-level representation from the text image. The RNN (LSTM) allows exploiting content information as a sequence labeling problem \cite{he2016reading}. In this study, two models are experimentally tested: the baseline method and the proposed method. The architecture of these models remains the same except an additional head is added at the end of the network for the auxiliary task in the case of the proposed method. As shown in Figure~\ref{model_architecture} the model comprises three major components: the CNN for automatic feature extraction, RNN for sequence learning, and CTC as output transcriber. In the proposed model, due to the promises of multi-task learning from the nature of the Amharic characters \cite{gondere2019handwritten}, the sum of the CTC losses for the character label and row label is optimized. To give emphasis to the difference between the competing models, basic configurations are set for CNN and RNN components as described in Figure~\ref{model_architecture}.

\begin{equation}
	CTC \: loss^{total}=CTC \: loss^{char} + CTC \: loss^{row}
	\label{eq1}
\end{equation}
\begin{equation}
	CTC \: loss=- \sum_{(x,l) \in D} \log \, p(l|x)
	\label{eq2}
\end{equation}
\begin{equation}
	p(l|x)=\sum_{\pi \in F^{-1}(l)} p(\pi|x)
	\label{eq3}
\end{equation}
\begin{equation}
	p(\pi|x)=\prod_{t=1}^{T}y_{\pi_t}^t
	\label{eq4}
\end{equation}

In this study, Eqs. \ref{eq1} to \ref{eq4} are used for defining the CTC loss. Eq. \ref{eq1} presents the total loss to be optimized which is implemented by the proposed method. It comprises the sum of the two CTC losses: CTC for character based transcription (baseline method), and CTC for row based transcription. The objective of the CTC loss as shown in Eq. \ref{eq2} is to minimize the negative log probability. \(p(l|x)\) is the conditional probability of the labels for a given training set \((D)\) consisting input \((x)\) and label \((l)\) sequences.The conditional probability \(p(l|x)\) of a given mapped label \(l\) is calculated using the sum of probabilities of all the alignment paths \(\pi\) mapped to that label using Eq. \ref{eq3}. The conditional probability \(p(\pi|x)\) of a given alignment path as shown in Eq. \ref{eq4} is calculated by the product of all labels including blank characters occurring at each time step along that path. Where \(\pi_t\) is the label occurred at time \(t\), \(y^t_{\pi_t}\) is probability of the label, and \(T\) is the input sequence length. It should be noted that paths should be mapped using a function \((F)\) into correct label sequences by removing the repeated symbols and blanks for providing the correct label path for the set of aligned paths. Further details about the baseline method can be found in \cite{abdurahman2021ahwr} and a broader insight of the proposed model is presented in \cite{gondere2019handwritten,gondere2021multiscript}.

\section{Experimental Results and Discusion }
All the experiments in this study are implemented using the Pytorch machine learning library on the Google Colab and the computing cluster of Information Systems and Machine Learning Lab (ISMLL) from the University of Hildesheim. Each character image is \(32\times32\) pixel and hence the word images through concatenation will have a height of \(32\) and width of \(32\) times the number of characters in that word. Accordingly, to address the variety of word lengths and maintain the natural appearance of the word images, models are trained using stochastic gradient descent.

Due to high computation costs, experiments are set to run for only fewer epochs by examining the learning behaviors. Hence during training, both the baseline and proposed models run for 30 epochs which is taking a training time of 4hrs. In each and several trials, a significant superiority was observed from the proposed model over the baseline. That is a largely better result is achieved by the proposed model even in the earlier training epochs. Hence, we let the baseline model run up to 30 more epochs to further examine the differences. Finally, the trained models from both setups were tested using a separate test set. The word error rate (WER) and character error rate (CER) are used as evaluation metrics. Further, the loss and accuracy curves of the models are presented to show the differences in the learning behavior.

\begin{figure}[!ht]
	\centering
	\includegraphics[width=0.75\textwidth]{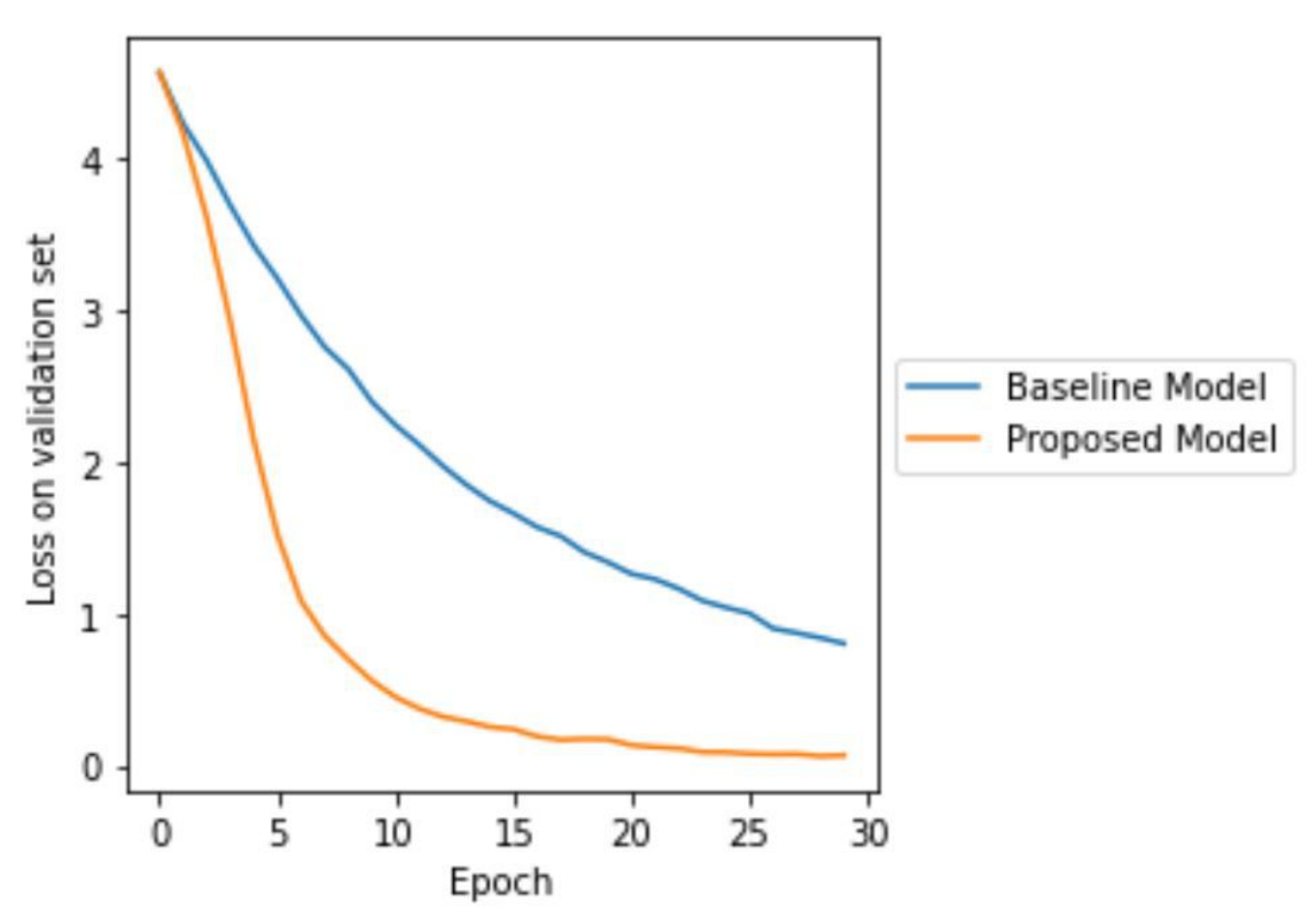}
	\caption{The loss curves of the models on validation set.} \label{lossval}
\end{figure}

\begin{figure}[!h]
	\centering
	\includegraphics[width=0.75\textwidth]{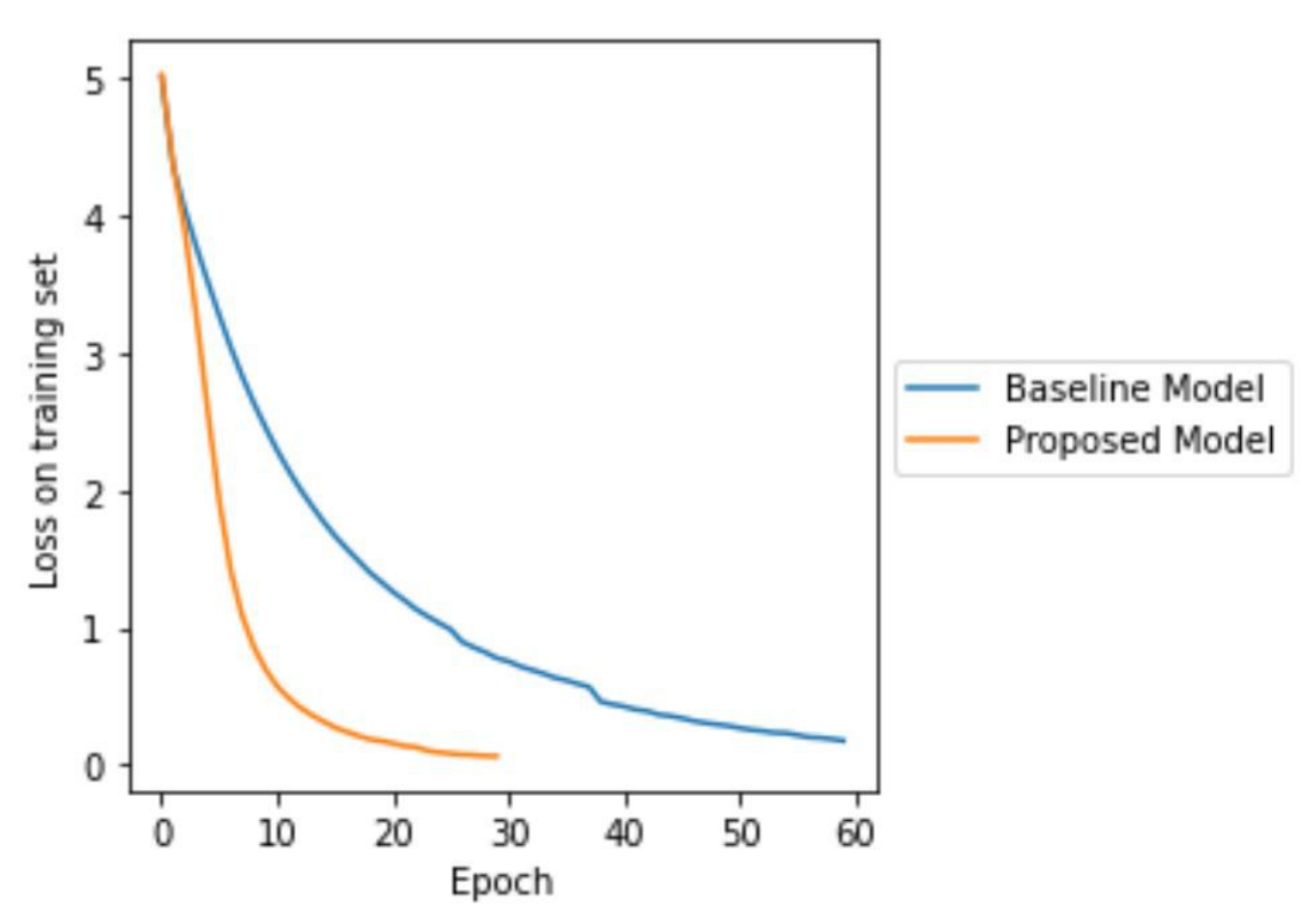}
	\caption{The learning behavior (loss curve) of the models on training set.} \label{losstrain}
\end{figure}

\begin{figure}[!t]
	\centering
	\includegraphics[width=0.75\textwidth]{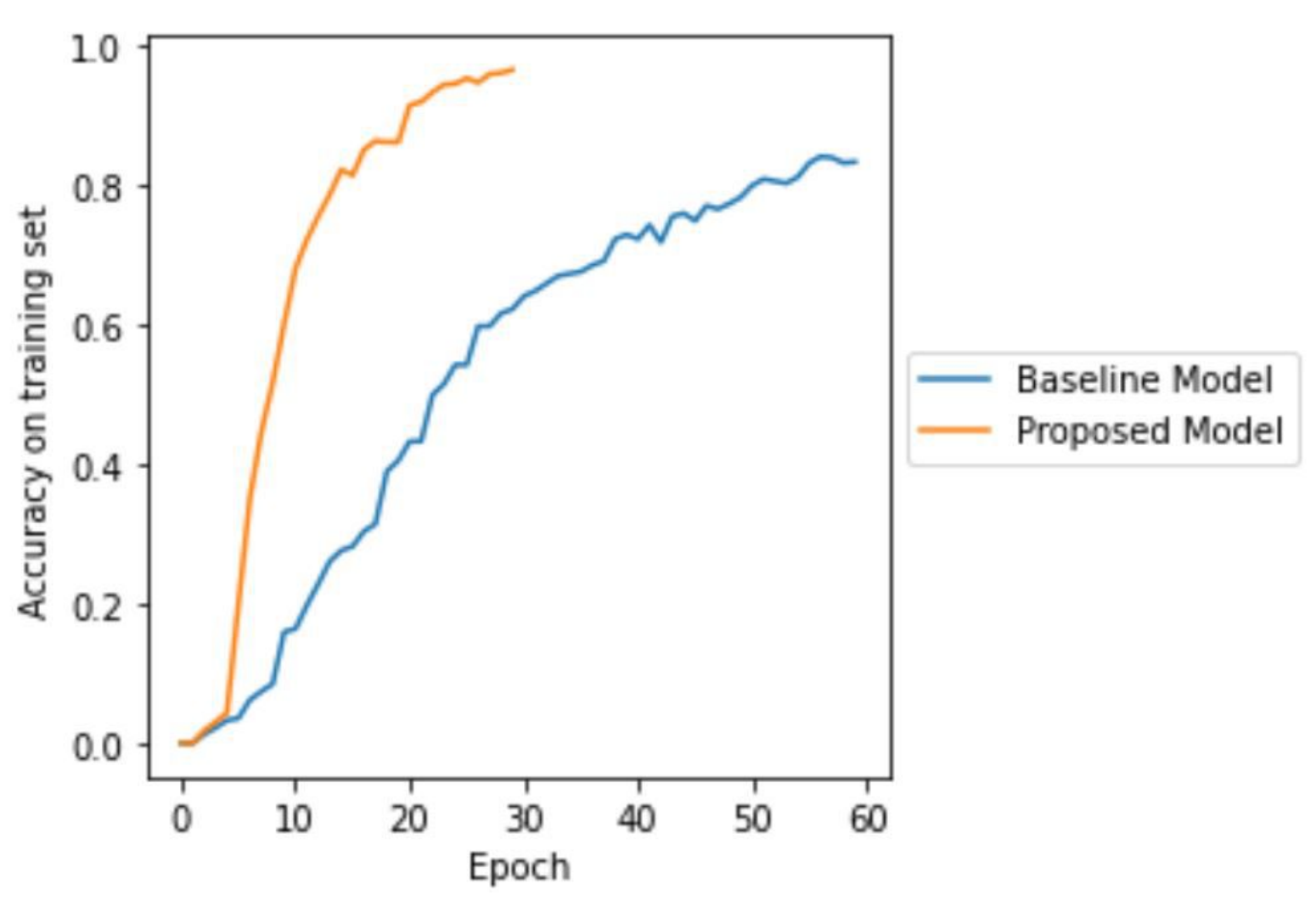}
	\caption{The learning behavior (accuracy curve) of the models on training set.} \label{acctrain}
\end{figure}

As can be seen in Figure~\ref{lossval}, the proposed model has converged much earlier since the 20\textsuperscript{th} epoch. The loss of the baseline method has reached 0.8035 at the 30\textsuperscript{th} epoch on the validation set. However, the loss of the proposed model reached 0.0664 at the 30\textsuperscript{th} epoch on the validation set. Similarly, the training behaviors of the models as shown in Figures~\ref{losstrain} and \ref{acctrain} demonstrate the superiority of the proposed model. The losses of the proposed and baseline models on the training set during the 30\textsuperscript{th} and 60\textsuperscript{th} epochs are 0.0638 and 0.1774 respectively. This result demonstrates how the proposed model quickly optimizes the loss during training. Likewise, the accuracy of the baseline model on the training set during the 60\textsuperscript{th} epoch (83.39\%) is still less than that of the proposed model (96.61\%) during 30\textsuperscript{th} epoch. As shown in Table~\ref{tab1} the WER and CER of the baseline model on the test set are 16.67 and 5.52 respectively. The proposed model on the other hand outperformed with WER of 4.21 and CER of 1.04 with only up to 30\textsuperscript{th} epoch trained model.

In this study, the significant supremacy of the proposed model can be another empirical evidence for the relevance of multi-task learning through the related tasks that emerged from the nature of the Amharic alphabet in Amharic handwritten character recognition which was explored in \cite{gondere2019handwritten}. More importantly, in this study which applied sequence based learning, the integration of multi-task learning that allowed the exploitation of the information contained within characters to complement the whole recognition performance is an interesting finding. Even though, the datasets and architectures used by other similar studies vary, the significant improvement by the proposed model over the baseline model in this study implies the importance of the proposed model. That is, while the baseline model recognition result is less than the results reported by other studies, the proposed model however has surpassed all of them as shown in Table~\ref{tab1}. The result reported by Abdurahman et al. \cite{abdurahman2021ahwr} is a WER of 5.24 and a CER of 1.15. The authors used the various CRNN configurations with CTC loss like the baseline model in this study. The authors have compiled 12,064 Amharic handwritten word images from 60 writers and have done manual augmentation of up to 33,672 handwritten Amharic word images. Belay et al. \cite{belay2020amharic} reported a CER of 1.59 for printed Amharic optical character recognition using a related architecture formulated in CRNN and CTC loss. From the total text-line images (337,337) in their dataset, 40,929 are printed text-line images written with the Power Geez font; 197,484 and 98,924 images are synthetic text-line images generated with different levels of degradation using the Power Geez and Visual Geez fonts, respectively.

\begin{table}[!t]
	\caption{Comparison of results on test set}\label{tab1}
	\centering
	\begin{tabular}{lp{3.5cm}lcc}
		\hline
		Model & Dataset & Method & WER (\%) & CER (\%)\\
		\hline
		Belay et al. \cite{belay2020amharic} & \raggedright337,337 printed Amharic text-lines & CRNN+CTC & - & 1.59\\
		
		Abdurahman et al. \cite{abdurahman2021ahwr} & \raggedright33,672 handwritten Amharic words& CRNN+CTC & 5.24 & 1.15\\
		
		Our Baseline & \raggedright256,100 constructed handwritten words & CRNN+CTC & 16.67 & 5.52\\
		
		Our Proposed & \raggedright256,100 constructed handwritten words & CRNN+2CTC & \textbf{4.21} & \textbf{1.04}\\
		\hline
	\end{tabular}
\end{table}

\section{Conclusion}
Several studies are emerging to address text image recognition as a sequence learning problem. This was possible due to the integration of convolutional neural networks, recurrent neural networks, and connectionist temporal classification. Hence, various works have addressed the OCR problem in an end-to-end fashion using CRNN and CTC loss. They have also demonstrated the suitability of this approach and the improved results. From the advantages of different deep learning innovations and opportunities gained in specific languages, it is important to propose novel approaches which can again further be generalized to different problems. Hence, in this study due to the nature of the Amharic alphabet organization allowing different parallel classification tasks and the advantages of multi-task learning, a new way of improving the traditional CRNN with CTC loss approach is proposed. The traditional CRNN with CTC loss model is implemented in an end-to-end fashion as a baseline model to address the problem of Amharic handwritten word recognition. The proposed model on the other hand is implemented by complementing the loss function using the auxiliary task from the row-wise similarities of the Amharic alphabet. The results of the study demonstrated significant improvements by the proposed model both in performance and learning behavior.

Finally, in this study, for a quick demonstration of the proposed model, only the row-wise similarity was explored as a related task which is very promising since characters in the same row in the Amharic alphabet have a very similar shape. However, a similar setup with column-wise similarity or integrating both can be explored as important future research work.

%
%
%
%

\end{document}